\documentclass[11pt,a4paper]{article}
\usepackage[hyperref]{acl2019}
\usepackage{times}
\usepackage{latexsym}
\usepackage{amsmath}
 \usepackage{amssymb}
\usepackage{graphicx}
\usepackage{booktabs}
\usepackage{algorithm}
\usepackage{algorithmic}
\usepackage{multirow}
\usepackage{amsfonts,amssymb}
\usepackage{amsmath,bm}
\usepackage{amsmath}
\usepackage{subfigure}

\usepackage{url}

\aclfinalcopy 

\title{Context-aware Embedding for Targeted Aspect-based Sentiment Analysis}

\author{Bin Liang$^{1}$, Jiachen Du$^{1}$, Ruifeng Xu$^{1}$\thanks{$^\ast$ Corresponding Author} , Binyang Li$^{2}$, Hejiao Huang$^{1}$\\
  $^{1}$Department of Computer Science, Harbin Institute of Technology (Shenzhen), China \\
  $^{2}$ School of Information Science and Technology, \\
  University of International Relations, Beijing, China \\
  {\tt 18b951033@stu.hit.edu.cn, dujiachen@stmail.hitsz.edu.cn}\\
  {\tt xuruifeng@hit.edu.cn, byli@uir.edu.cn}\\
  {\tt huanghejiao@hit.edu.cn}
  }

\date{}

\begin{document}
\maketitle
\begin{abstract}
Attention-based neural models were employed to detect the different aspects and sentiment polarities of the same target in targeted aspect-based sentiment analysis (TABSA). However, existing methods do not specifically pre-train reasonable embeddings for targets and aspects in TABSA. This may result in targets or aspects having the same vector representations in different contexts and losing the context-dependent information. To address this problem, we propose a novel method to refine the embeddings of targets and aspects. Such pivotal embedding refinement utilizes a sparse coefficient vector to adjust the embeddings of target and aspect from the context. Hence the embeddings of targets and aspects can be refined from the highly correlative words instead of using context-independent or randomly initialized vectors. Experiment results on two benchmark datasets show that our approach yields the state-of-the-art performance in TABSA task.
\end{abstract}

\section{Introduction}

Targeted aspect-based sentiment analysis (TABSA) aims at detecting aspects according to the specific target and inferring sentiment polarities corresponding to different target-aspect pairs simultaneously~\cite{DBLP:conf/coling/SaeidiBLR16}. For example, in sentence ``{\em location1 is your best bet for secure although expensive and location2 is too far.}", for target ``location1", the sentiment polarity is positive towards aspect ``{\em SAFETY}" but is negative towards aspect ``{\em PRICE}". While ``location2" only express negative polarity about aspect ``{\em TRANSIT-LOCATION}". This can be seen in Figure 1, e.g., where opinions on the aspects ``{\em SAFETY}" and ``{\em PRICE}" are expressed for target ``{\em location1}" but not for target ``{\em location2}", whose corresponding aspect is ``{\em TRANSIT-LOCATION }". Here, an interesting phenomenon is that, the opinion ``Positive" towards aspect ``{\em SAFETY}" is expressed for target ``{\bf location1}" will be change if ``{\bf location1}" and ``{\bf location2}" are exchanged. That is to say, the representation of target and aspect should take full account of context information rather than use context-independent representation.

\begin{figure}[!htp]
\centering
\includegraphics[width=0.45\textwidth]{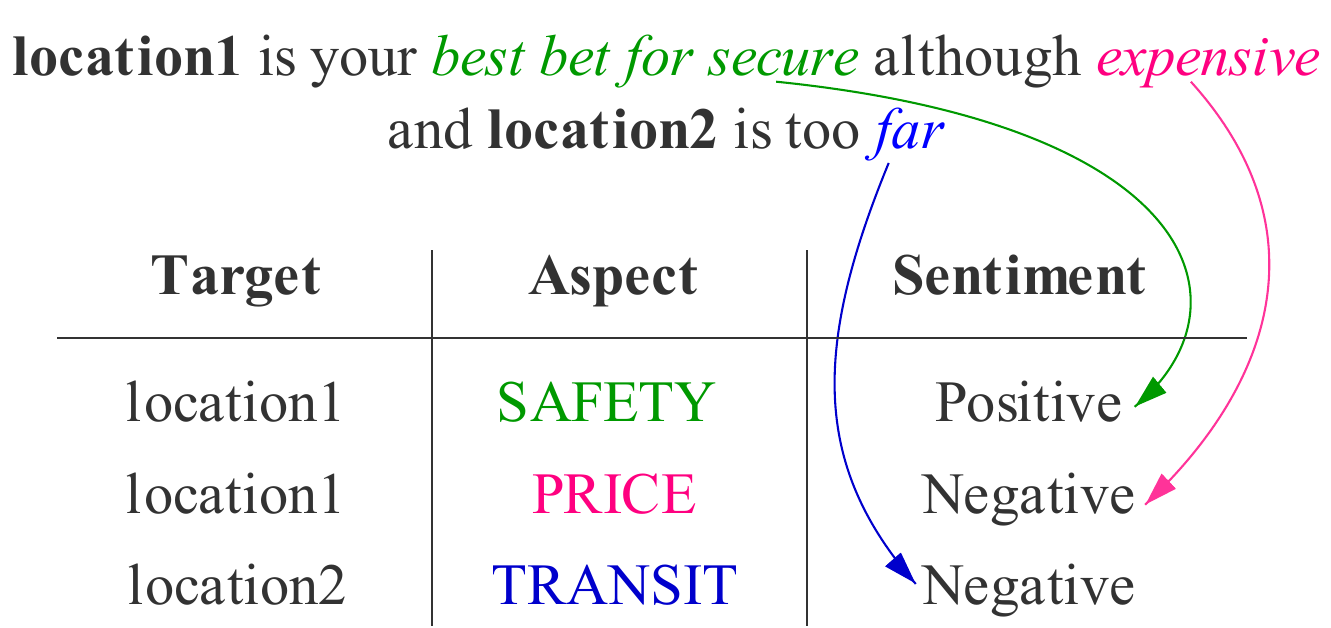}
\caption{Example of TABSA task. Highly correlative words and corresponding aspects are in the same color. Entity names are masked by {\bf location1} and {\bf location2}.} 
\end{figure}

Aspect-based sentiment analysis (ABSA) is a basic subtask of TABSA, which aims at inferring the sentiment polarities of different aspects in the sentence~\cite{ruder-ghaffari-breslin:2016:EMNLP2016,chen-EtAl:2017:EMNLP20171,Gui-EtAl:Learning,P18-1233,ma-etal-2018-joint}. Recently, attention-based neural models achieve remarkable success in ABSA~\cite{DBLP:conf/sigir/FanGD0XW18,wang-EtAl:2016:EMNLP20163,tang-qin-liu:2016:EMNLP2016}. In TABSA task, the attention-based sentiment LSTM~\cite{DBLP:conf/aaai/MaPC18} is proposed to tackle the challenges of both aspect-based sentiment analysis and targeted sentiment analysis by incorporating external knowledge. For neural model improvement, a delayed memory is proposed to track and update the states of targets at the right time with external memory~\cite{Liu+:2018}. 

Despite the remarkable progress made by the previous works, they usually utilize context-independent or randomly initialized vectors to represent targets and aspects, which losses the semantic information and ignores the interdependence among specific target, corresponding aspects and the context. Because the targets themselves have no expression of sentiment, and the opinions of the given sentence are generally composed of words highly correlative to the targets. For example, if the words ``price" and ``expensive" are in the sentence, it probably expresses the ``Negative" sentiment polarity about aspect ``{\em PRICE}".

To alleviate these problems above, we propose a novel embedding refinement method to obtain context-aware embedding for TABSA. Specifically, we use a sparse coefficient vector to select highly correlated words from the sentence, and then adjust the representations of target and aspect to make them more valuable. The main contributions of our work can be summarized as follows:
\begin{itemize} 
\item We reconstruct the vector representation for target from the context by means of a sparse coefficient vector, hence the representation of target is generated from highly correlative words rather than using context-independent or randomly initialized embedding. 
\item The aspect embedding is fine-tuned to be close to the highly correlated target and be away from the irrelevant targets. 
\item Experiment results on SentiHood and Semeval 2015 show that our proposed method can be directly incorporated into embedding-based TABSA models and achieve state-of-the-art performance.
\end{itemize}

\section{Methodology}
In this section, we describe the proposed method in detail. The framework of our proposed method is demonstrated in Figure 2. 

We assume a words sequence of a given sentence as an embedding matrix ${{\bf{X}}} \in {\mathbb R^{m \times n}}$, where ${n}$ is the length of sentence, ${m}$ is the dimension of embedding, and each word can be represented as an $m$-dimensional embedding ${\bf{x}} \in \mathbb {R} {^m}$ including the embedding of target ${\bf{t}} \in \mathbb {R} {^m}$ via random initialization and the embedding of aspect ${\bf{a}} \in \mathbb {R} {^m}$ which is an average of its constituting word embeddings or single word embedding. The sentence embedding matrix ${{\bf{X}}}$ is fed as input into our model to achieve the sparse coefficient vector $\bf u'$ via the fully connected layer and the step function successively. The hidden output $\bf u'$ is utilized to compute the refined representation of target ${{\tilde{\bf t}}\in \mathbb {R}^m}$ and aspect ${{\tilde{\bf a}}\in \mathbb {R}^m}$. Afterwards, the squared Euclidean function $d({\tilde{\bf t}},{\bf {t}})$ and $d({\tilde{\bf a}},{\tilde {\bf t}},{\bf {t}'})$ are used to iteratively minimize the distance to get the refined embeddings of target and aspect.

\begin{figure}[!htp]
\centering
\includegraphics[width=0.45\textwidth]{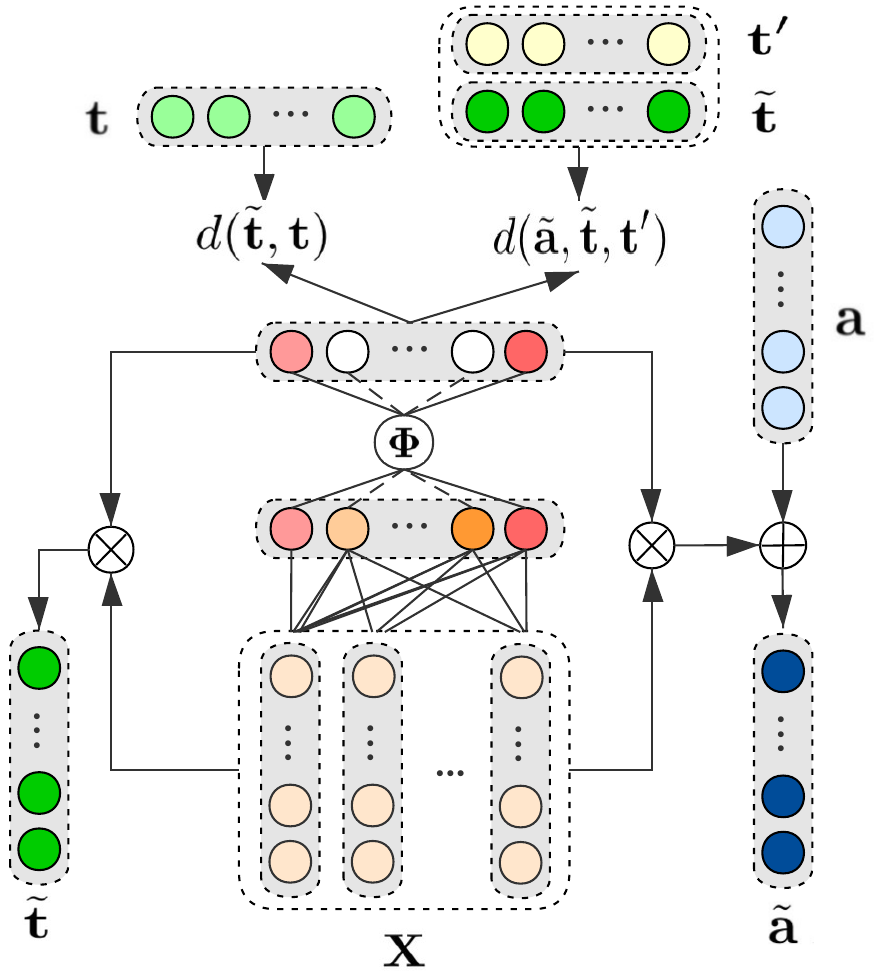}
\caption{The framework of our refinement model. $\otimes$ is element-wise product, $\oplus$ is vector addition, $\Phi$ is step function.} 
\end{figure}

\subsection{Task Definition}
Given a sentence consisting of a sequence of words $s = \{ {w_1},{w_2}, \ldots ,{LOC}, \ldots ,{w_n}\} $, where $LOC$ is a target in the sentence, there will be 1 or 2 targets in the sentence corresponding to several aspects. There are a pre-identified set of targets $T$ and a fixed set of aspects $A$. The goal of TABSA can be regarded as a fine-grained sentiment expression as a tuple ${(t, a, p)}$, where $p$ refers to the polarity which is associated with aspect $a$, and the aspect $a$ belongs to a target $t$. The objective of TABSA task is to detect the aspect $a \in A$ and classify the sentiment polarity $p \in \{Positive, Negative, None\}$ according to a specific target $t$ and the sentence $s$.

\subsection{Target Representation}
The idea of target representation is to reconstruct the target embedding from a given sentence according to the highly correlated words in the context. By this means we can extract the correlation between target and context, the target representation is computed as: 
\begin{equation}
{\tilde {\bf t} }={\bf X} \ast \bf u'
\end{equation}
where ${\tilde {\bf t} }$ is the representation of target, $\bf u'$ is a sparse coefficient vector indicating the importance of different words in the context, defined as:
\begin{equation}
{\bf u'=\Phi(\bf u)}
\end{equation}
where $\Phi$ is a step function given a real value:
\begin{equation}
\Phi(u_i)=
\begin{cases}
u_i& {u_i \geqslant mean(u)}\\
0& {u_i<mean(u)}
\end{cases}
\end{equation}
where $mean(\cdot)$ is an average function, and the vector {\bf u} can be computed by a non-linear function of basic embedding matrix ${\bf X}$:
\begin{equation}
{\bf u}=f({{\bf X}^\intercal}\cdot{\bf{W}}+{\bf b})
\end{equation}
where $f$ is a non-linear operation function like sigmoid, ${\bf{W}} \in \mathbb {R} {^{m}}$ and ${\bf b} \in \mathbb {R}^n$ denote the weight matrix and bias respectively. The target representation is inspired by the recent success of embedding refinement~\cite{yu-etal-2017-refining}. For each target, our reconstruction operation aims to get a contextual relevant embedding by iteratively minimizes the squared Euclidean between the target and the highly correlative words in the sentence. The objective function is defined as:
\begin{equation}
d({\tilde{\bf t}},{\bf {t}}) =\sum_{i=1}^{n}\Big({\sum_{j=1}^{m}({{\tilde {t} }_i^j-t_i^j})^2 + \lambda {u'_i}}\Big )
\end{equation}
where $\lambda {u'_i}$ aims to control the sparseness of vector ${\bf u'}$. Through the iterative procedure, the vector representation of the target will be iteratively updated until the number of the non-zero elements of vector $\bf u'$ less than the threshold value: $k \leqslant c$. Where $k$ is the number of the non-zero elements of vector $\bf u'$ and $c$ is a threshold to control the non-zero number of vector $\bf u'$.

\subsection{Aspect Representation}
Generally, the words of aspects contain the most important semantic information. Coordinate with the aspect itself, the context information can also reflect the aspect, such as the word ``price" in the sentence probably has relevant to aspect ``{\em PRICE}". To this end, we refine the aspect representation according to target representation. By incorporating highly correlated words into the representation of aspect, every element in the fine-tuned aspect embedding ${\tilde {\bf a}}$ is calculated as:
\begin{equation}
{{\tilde a}_i}={{a}_i}+\alpha{{X}_i} \ast {u}'_i
\end{equation}
where $\alpha$ is a parameter to control the influence between aspect and the context. 

For each aspect, the fine-tuning method aims to move closer to the homologous target and further away from the irrelevant one by iteratively minimizes the squared Euclidean. The objective function is thus divided into two parts:

\begin{small}
\begin{equation}
d({\tilde{\bf a}},{\tilde {\bf t}},{\bf {t}'})=\sum_{i=1}^{n}\Big[{\sum_{j=1}^{m}\Big(({{\tilde {a} }_i^j-{\tilde t}_i^j})^2-\beta({{\tilde {a} }_i^j-{t'}_i^j})^2 \Big) + \lambda {u'_i}}\Big] 
\end{equation}
\end{small}where $\bf{\tilde t}$ is the the homologous target and $\bf{t}'$ is the irrelevant one. $\beta$ is a parameter that controls the distance from the irrelevant target.

\begin{table*}[!htp]
\centering
\label{tab:booktabs}
\begin{tabular}{lccccc}
\toprule
\multirow{2}*{Model} & \multicolumn{3}{c}{Aspect Detection} & \multicolumn{2}{c}{Sentiment Classification} \\
~ & Acc. (\%) & F1 (\%) & AUC (\%) & Acc. (\%) & AUC (\%) \\
\midrule
LSTM-Final & --- & 68.9 & 89.8 & 82.0 & 85.4 \\
LSTM-Loc & --- & 69.3 & 89.7 & 81.9 & 83.9 \\
\midrule
SenticLSTM & 67.4 & 78.2 & --- & 89.3 & --- \\
{\bf RE}+SenticLSTM$\dagger$ {\bf (ours)} & 73.8 & 79.3 & --- & {\bf 93.0} & --- \\
\midrule
Delayed-memory & 73.5 & 78.5 & 94.4 & 91.0 & 94.8 \\
{\bf RE}+Delayed-memory$\dagger$ {\bf (ours)}& {\bf 76.4} & {\bf 81.0} & {\bf 96.8} & 92.8 & {\bf 96.2} \\
\bottomrule
\end{tabular}
\caption{Experimental results on SentiHood. $\dagger$ denotes average score over 10 runs, and best scores are in bold.}
\end{table*}

\begin{table*}[!h]
\centering
\label{tab:booktabs}
\begin{tabular}{lccccc}
\toprule
\multirow{2}*{Model} & \multicolumn{3}{c}{Aspect Detection} & \multicolumn{2}{c}{Sentiment Classification} \\
~ & Acc. (\%) & F1 (\%) & AUC (\%) & Acc. (\%) & AUC (\%) \\
\midrule
SenticLSTM & 67.3 & 76.4 & --- & 76.5 & --- \\
{\bf RE}+SenticLSTM$\dagger${\bf (ours)} & 71.2 & 78.6 & 89.5 & 76.8 & 82.3 \\
\midrule
Delayed-memory & 70.3 & 77.4 & 90.8 & 76.4 & 83.6\\
{\bf RE}+Delayed-memory$\dagger${\bf (ours)} & {\bf 71.6} & {\bf 79.1} & {\bf 91.8} & {\bf 77.2} & {\bf 84.6} \\
\bottomrule
\end{tabular}
\caption{Experimental results on Semeval 2015.}
\end{table*}

\section{Experiments}
This section evaluates several deep neural models based on our proposed embedding refinement method for TABSA.

\paragraph{Dataset.} Two benchmark datasets: SentiHood~\cite{DBLP:conf/coling/SaeidiBLR16} and Task 12 of Semeval 2015~\cite{pontiki-EtAl:2015:SemEval} are used to evaluate our proposed method. SentiHood contains annotated sentences containing one or two location target mentions. The whole dataset contains 5215 sentences with 3862 sentences containing a single location and 1353 sentences containing multiple (two) locations. Location target names are masked by $\bf LOCATION1$ and $\bf LOCATION2$ in the whole dataset. Following ~\cite{DBLP:conf/coling/SaeidiBLR16}, we only consider the top 4 aspects (``GENERAL", ``PRICE", ``TRANSIT-LOCATION" and ``SAFETY") when evaluate aspect detection and sentiment classification. To show the generalizability of our method, we evaluate our works in another dataset: restaurants domain in Task 12 for TABSA from Semeval 2015. We remove sentences containing no targets as well as {\em NULL} targets like the work of~\cite{DBLP:conf/aaai/MaPC18}. The whole dataset contains 1,197 targets in the training set and 542 targets in the testing set.

\paragraph{Experiment setting.} We use Glove~\cite{pennington-socher-manning:2014:EMNLP2014}\footnote{http://nlp.stanford.edu/projects/glove/} to initialize the word embeddings in our experiments, and target embeddings (location1 and location2) are randomly initialized. We initialize $\bf W$ and $\bf b$ with random initialization. The parameters of $c$, $\alpha$ and $\beta$ in our experiment are 4, 1 and 0.5 respectively. Given a unit of text $s$, a list of labels $(t, a, p)$ is provided correspondingly, the overall task of TABSA can be defined as a three-class classification task for each $(t, a)$ pair with labels {\em Positive}, {\em Negative}, {\em None}. We use macro-average {\em F1}, Strict accuracy (Acc.) and AUC for aspect detection, and Acc. and AUC for sentiment classification ignoring the class of {\em None}, which indicates that a sentence does not contain an opinion for the target-aspect pair $(t, a)$.

\paragraph{Comparison methods.} We compare our method with several typical baseline systems~\cite{DBLP:conf/coling/SaeidiBLR16} and remarkable models~\cite{DBLP:conf/aaai/MaPC18,Liu+:2018} proposed for the task of TABSA. 

(1) {\bf LSTM-Final}~\cite{DBLP:conf/coling/SaeidiBLR16}: A bidirectional LSTM model takes the final states to represent the information.

(2) {\bf LSTM-Loc}~\cite{DBLP:conf/coling/SaeidiBLR16}: A bidirectional LSTM model takes the output representation at the index corresponding to the location target.

(3) {\bf SenticLSTM}~\cite{DBLP:conf/aaai/MaPC18}: A bidirectional LSTM model incorporates external SenticNet knowledge.

(4) {\bf Delayed-memory}~\cite{Liu+:2018}: A memory-based model utilizes a delayed memory mechanism.

(5) {\bf RE+SenticLSTM}: Incorporating our proposed method into SenticLSTM.

(6) {\bf RE+Delayed-memory}: Incorporating our proposed method into Delayed-memory.

\subsection{Comparative Results of SentiHood}
The experimental results are shown in Table 1. The classifiers based on our proposed methods ({\bf RE}+Delayed-memory, {\bf RE}+SenticLSTM) achieve better performance than competitor models for both aspect detection and sentiment classification. In comparison with the previous best-performing model (Delayed-memory), our best model ({\bf RE}+Delayed-memory) significantly improves aspect detection (by 2.9\% in strict accuracy, 2.5\% in macro-average F1 and 2.4\% in AUC) and sentiment classification (by 1.8\% in strict accuracy and 1.4\% in AUC) on SentiHood.

The comprehensive results show that by incorporating refined context-aware embeddings of targets and aspects into the neural models can substantially improve the performance of aspect detection. This indicates that the refined representation is more learnable and is able to extract the interdependence between aspect and the corresponding target in the context. On the other hand, the performance of sentiment classification is improved certainly in comparison with the remarkable models (Delayed-memory and SenticLSTM). It indicates that our context-aware embeddings can capture sentiment information better than the models using traditional embeddings even incorporating external knowledge.

\subsection{Comparative Results of Semeval 2015}
To illustrate the robustness of our proposed method, a comparative experiment was conducted on Semeval 2015. As shown in Table 2, our embedding refinement method achieves better performance for both aspect detection and sentiment classification than two original embedding-based models, for aspect detection in particular. Consequently, our method is capable of achieving state-of-the-art performance on different datasets.

\begin{figure}[!htp]
\centering
\subfigure[{\bf RE}+Delayed-memory.]{
\begin{minipage}[b]{0.225\textwidth}
\includegraphics[width=1\textwidth]{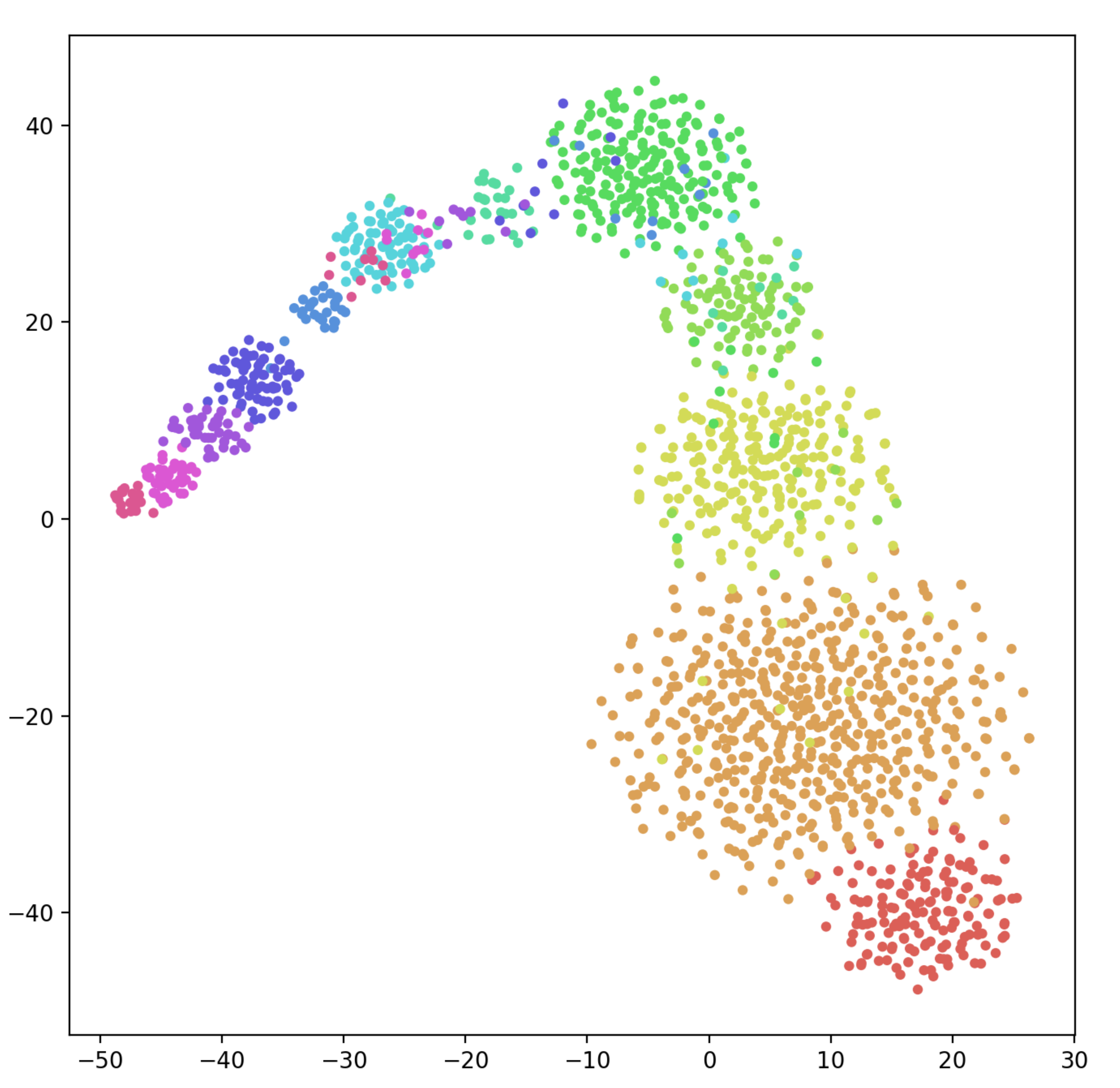} \\
\end{minipage}
}
\subfigure[Delayed-memory.]{
\begin{minipage}[b]{0.225\textwidth}
\includegraphics[width=1\textwidth]{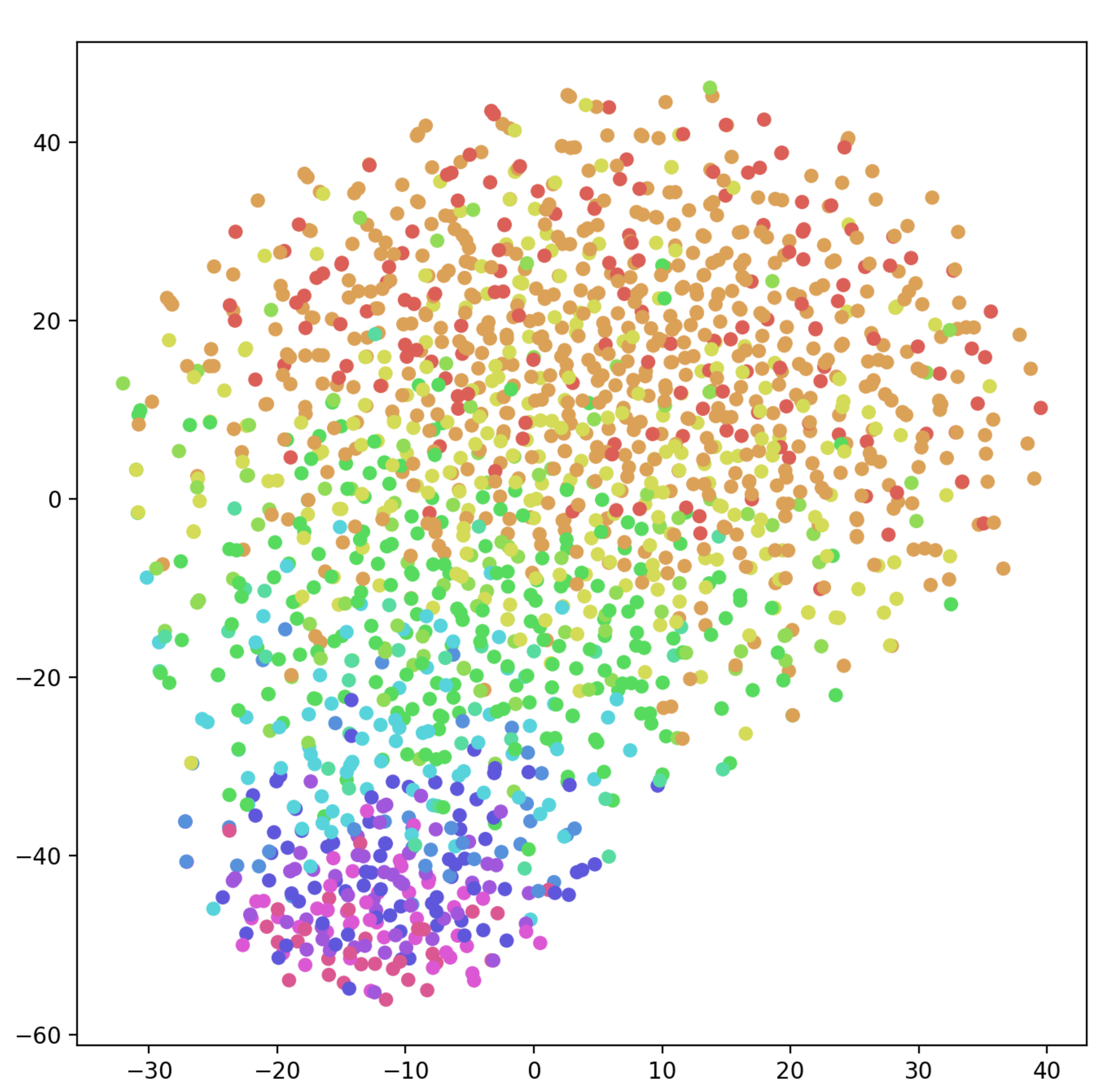} \\
\end{minipage}
}
\subfigure[{\bf RE}+SenticLSTM.]{
\begin{minipage}[b]{0.225\textwidth}
\includegraphics[width=1\textwidth]{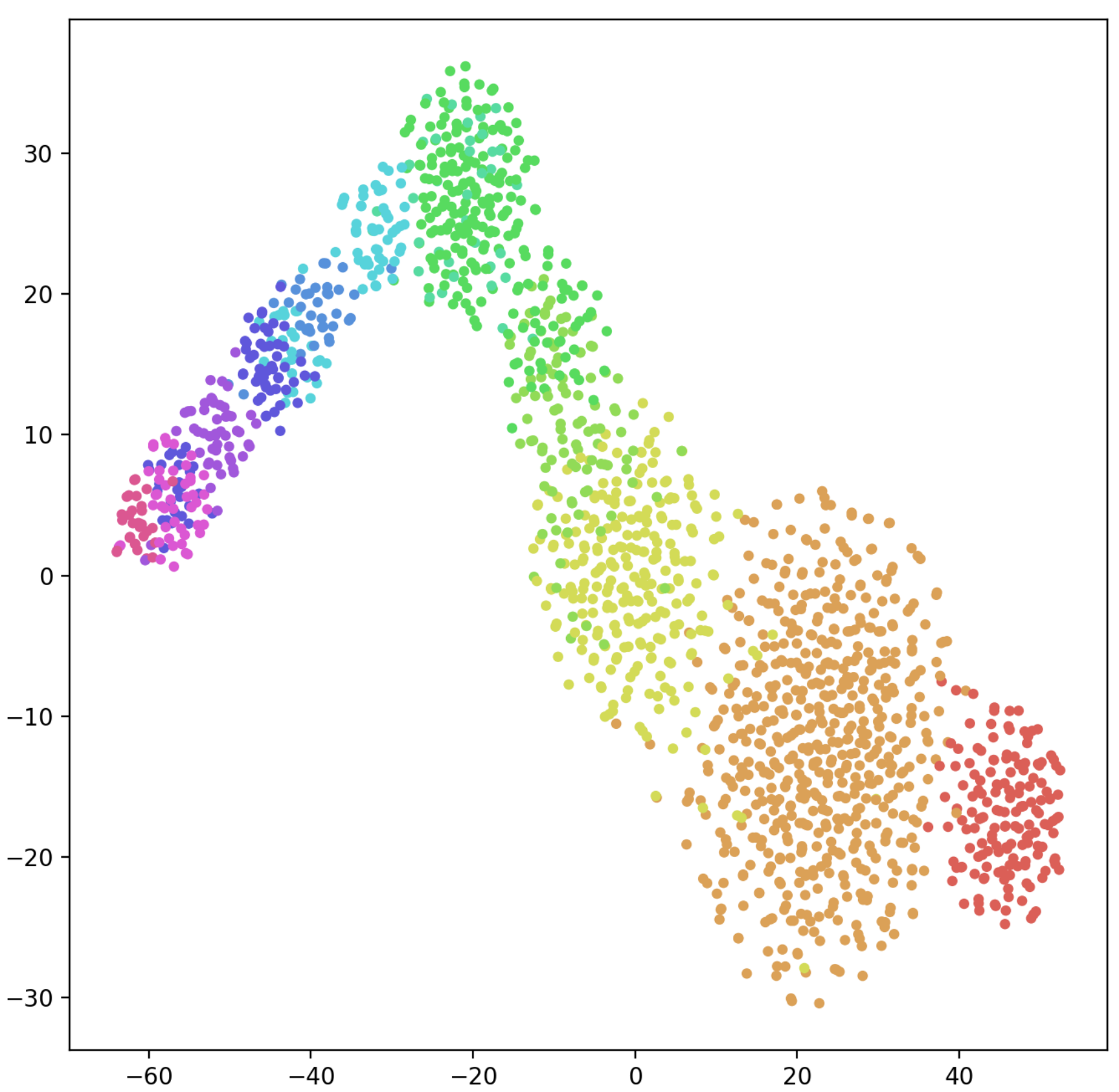} \\
\end{minipage}
}
\subfigure[SenticLSTM.]{
\begin{minipage}[b]{0.225\textwidth}
\includegraphics[width=1\textwidth]{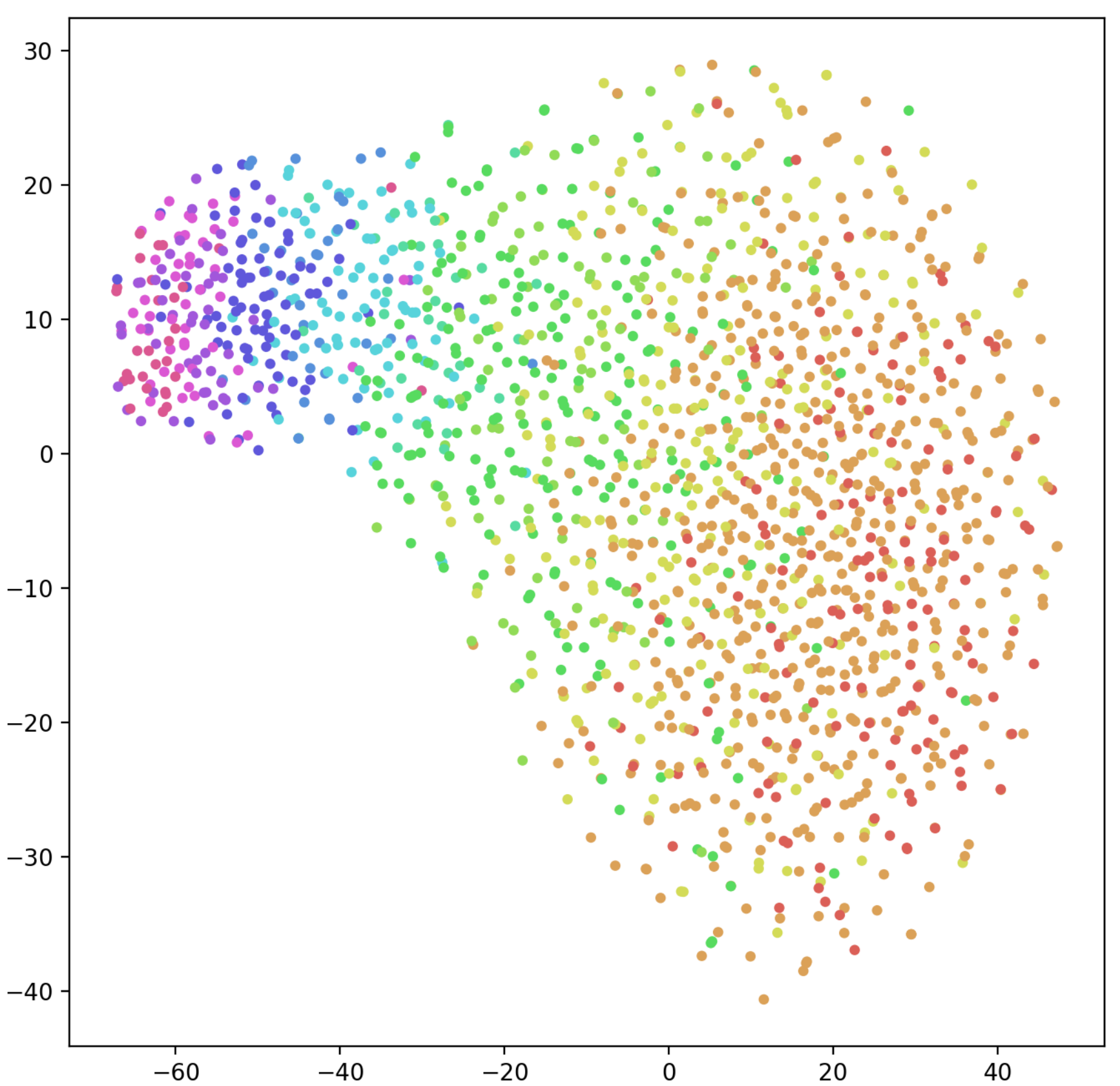} \\
\end{minipage}
}
\caption{The visualization of intermediate embeddings learned by embedding-based models. Different colors represent different aspects.} 
\end{figure}

\subsection{Visualization}
To qualitatively demonstrate how the proposed embedding refinement improves the performance for both aspect detection and sentiment classification in TABSA, we visualize the proposed context-aware aspect embeddings $\tilde{\bf a}$ and original aspect embeddings ${\bf a}$ which are learned with Delayed-memory and SenticLSTM models via t-SNE \cite{maaten2008visualizing}. As shown in Figure 3, compared with randomly initialized embedding, it is observed a significantly clearer separation between different aspects represented by our proposed context-aware embedding. This indicates that different representations of aspects can be distinguished from the context, and that the commonality of a specific aspect can also be effectively preserved. Hence the model can extract different semantic information according to different aspects, when detecting multiple aspects in the same sentence in particular. The results verify that encoding aspect by leveraging context information is more effective for aspect detection and sentiment classification in TABSA task. 

\section{Conclusion}

In this paper, we proposed a novel method for refining representations of targets and aspects. The proposed method is able to select a set of highly correlated words from the context via a sparse coefficient vector and then adjust the representations of targets and aspects. Hence, the interdependence among specific target, corresponding aspect, and the context can be extracted to generate superior embedding. Experimental results demonstrated the effectiveness and robustness of the proposed method on two benchmark datasets over the task of TABSA. In future works, we will explore the extension of this approach for other tasks.

\section*{Acknowledgments}
This work was supported by National Natural Science Foundation of China U1636103, 61632011, 61876053, U1536207, Key Technologies Research and Development Program of Shenzhen JSGG20170817140856618, Shenzhen Foundational Research Funding JCYJ20180507183527919, Joint Research Program of Shenzhen Securities Information Co., Ltd. No. JRPSSIC2018001.

\bibliography{acl2019}
\bibliographystyle{acl_natbib}

\end{document}